# A Study on Neuro-Symbolic Artificial Intelligence: Healthcare Perspectives

Delower Hossain[1,3], Jake Y Chen * [1, 2, 3]

***Abstract***— Over the last few decades, Artificial Intelligence (AI) scientists have been conducting investigations to attain human-level performance by a machine in accomplishing a cognitive task. Within machine learning, the ultimate aspiration is to attain Artificial General Intelligence (AGI) through a machine. This pursuit has led to the exploration of two distinct AI paradigms. Symbolic AI, also known as classical or GOFAI (Good Old-Fashioned AI) and Connectionist (Sub-symbolic) AI, represented by Neural Systems, are two mutually exclusive paradigms. Symbolic AI excels in reasoning, explainability, and knowledge representation but faces challenges in processing complex real-world data with noise. Conversely, deep learning (Black-Box systems) research breakthroughs in neural networks are notable, yet they lack reasoning and interpretability. Neuro-symbolic AI (NeSy), an emerging area of AI research, attempts to bridge this gap by integrating logical reasoning into neural networks, enabling them to learn and reason with symbolic representations. While a long path, this strategy has made significant progress towards achieving common sense reasoning by systems. This article conducts an extensive review of over 977 studies from prominent scientific databases (DBLP, ACL, IEEExplore, Scopus, PubMed, ICML, ICLR), thoroughly examining the multifaceted capabilities of Neuro-Symbolic AI, with a particular focus on its healthcare applications, particularly in drug discovery, and Protein engineering research. The survey addresses vital themes, including reasoning, explainability, integration strategies, 41 healthcare-related use cases, benchmarking, datasets, current approach limitations from both healthcare and broader perspectives, and proposed novel approaches for future experiments. Moreover, it identifies critical challenges in applying Neuro-Symbolic AI to medicine and represents a comprehensive exploration of its transformative potential in the biomedical field.

## I. Introduction

AI is an interdisciplinary field integrating cognitive science, philosophy, psychology, computer science, neuroscience, and other domains. At its core, modern AI is anticipated by deep learning, the main component of which is artificial neural networks (inspired by neurons workings of the human brain) [1]. Geoffrey Hinton, along with David E. Rumelhart et al., published an article in 1986 titled "Learning representations by back-propagating errors," which is widely recognized for reformulating and popularizing the backpropagation algorithm for training multilayer neural networks. He was awarded the 2024 Nobel Prize in Physics [2]. Autonomous vehicles, galaxies, and stars classification became a reality by leveraging this technology. However, despite AI's breakthroughs, interpretability, explainability, and reasoning remain opacity—critical aspects for secure applications such as in healthcare and autonomous systems. Neuro-symbolic AI (NeSy) addresses this gap by blending neural networks with symbolic logic, facilitating explainable reasoning, interpretability, and scalability in AI. To recognize how NeSy bridges the divide, it is essential to explore the foundations of AI, which are rooted in two paradigms: Symbolism and connectionism.

Symbolic AI, often referred to as Good Old-Fashioned Artificial Intelligence (GOFAI), represents a classical approach focusing on knowledge representation. This paradigm dominated the classical Artificial Intelligence era from the 1950s to the 1980s [3]. A defining characteristic of symbolic methods is their capacity for explanation and reasoning, facilitating decision-making processes. Symbolic techniques encompass explicit symbolic methods such as programming languages, first-order logic, propositional logic, and domain-specific symbolic expressions. Other procedures include rules, knowledge graphs, ontologies, decision trees, and symbolic

✉ Delower Hossain
mhossai5@uab.edu

✉ Jake Y Chen
jakechen@uab.edu

1 Dept. of Computer Science at The University of Alabama at Birmingham, USA

2 Dept. of Genetic, Biomedical Engineering, Informatics and Data Science, Heersink School of Medicine at The University of Alabama at Birmingham, USA

3 Systems Pharmacology AI Research Center, at The University of Alabama at Birmingham, USA

expressions. However, a notable limitation of this approach is its inability to manage large-scale and noisy data effectively.

Sub-symbolic AI is an approach to Artificial Intelligence distinct from Symbolic AI, often described as a “black box.” This black-box system draws on large-scale data, typically for statistical learning. Its development accelerated with the pioneering work of "godfathers" of deep learning [4]—researchers such as Geoffrey Hinton, Yann LeCun, and Yoshua Bengio—who contributed to fundamental advancements in connectionist approaches. An essential building block of Sub-Symbolic AI is the perceptron, developed by Frank Rosenblatt in the 1950s, while McCulloch and Pitts developed the early neuron model in 1943 [3]. The perceptron is a type of artificial neuron. Multilayer Perceptrons (MLPs) [5] and Deeper Neural Networks have since evolved, greatly extending the capabilities of Sub-Symbolic AI to handle complex, nonlinear problems. While Sub-Symbolic AI excels at pattern recognition and can hold vast amounts of unstructured and noisy data, it often lacks explainability and reasoning. This lack of transparency has led to the label of a "black box" approach, as its complex internal computations are typically opaque to users, making it challenging to interpret how decisions are made. Nevertheless, Sub-Symbolic AI remains integral to advancements in fields such as autonomous systems and healthcare due to its incapability and reasoning. As such, a new dimension of AI has captured the interest of researchers over the past decades.

Neuro-symbolic AI is a novel branch of Artificial Intelligence that aims to overcome the limitations of both symbolic and connectionist approaches by integrating their strengths. This integration seeks to develop AI systems that are not only explainable and interpretable but also capable of reasoning. In the 1990s, Geoffrey G. Towell presented new AI approaches recognized as Knowledge-Based Artificial Neural Networks (KBANN) (1994)[6] and NofM (1991)[7]. Recent research in Neuro-symbolic AI has yielded cognitive simulations and frameworks for learning, reasoning, and language, such as NS-VQA [8], achieving 99.8% accuracy on the CLEVR dataset. Numerous models and frameworks, such as; LTN [9], LNN [10], DFOL-VQA [11], FSKBANN [12], MultiPredGO [13], DeepMiR2GO [14], ExplainDR [15], PP-DKL [16], FSD [17], CORGI [18], KGCNN [19], NeurASP [20], have been developed during this period, showing promising performance in the biomedical and other sectors. This study examines the role of Neuro-symbolic AI (NeSy) in healthcare settings, focusing on its applications in biomedical use cases, model performance, implications for medical datasets, and emerging trends while evaluating state-of-the-art NeSy algorithms. Additionally, it offers simulation insights into drug discovery and presents a novel approach to advancing the protein-drug interaction use case task.

The article presents a comprehensive survey, evaluation, and analysis of the promising Neuro-Symbolic (NeSy) paradigm. The enhanced outline of the article is as follows: Section II outlines the study's process and methodology, followed by an overview of Neuro-symbolic AI, its categorizations and principles, and reviews contemporary models and approaches, including their results in Section III. Section IV focuses on NeSy applications in healthcare, while Section V addresses open challenges and limitations. Section VI proposes a new method and discusses future outlooks, with the article concluding in Section VII. To our knowledge, no prior survey or systematic review has explicitly focused on Neuro-Symbolic AI in healthcare perspectives, although several reviews in other domains have observed [21]- [43].

*Key contributions:*

- An overview of real-life application exploration of Neuro-Symbolic AI in Healthcare.
- Proposed novel approaches using large chemical-protein language models and Logic Tensor Networks for compound-protein interaction tasks.
- Comprehensive analysis of recent neuro-symbolic domains of applications uncovered promising models for benchmarking, dataset, model evaluation, healthcare, and the no-healthcare perspective.
- Revealed practical simulation results of NeSy approaches in drug discovery (Cardiotoxicity, Diabetes, and TNBC compound classification).
- Summarized Existing model strengths (Reasoning, Interpretability) and limitations.
- Identified open challenges in applying Neuro-symbolic AI to Biomedical field.

## II. METHODOLOGY

This section serves research questions, article selection criteria, and data extraction methods applied to studies across healthcare and non-healthcare domains.

*Research Questions*

**RQ1:** What are the recent advancements in Neuro-symbolic AI techniques and methodologies in healthcare and non-healthcare settings?

**RQ2:** What are the emerging trends and potential Applications of Neuro-Symbolic AI in the biomedical domain?

**RQ3:** What are the benefits of integrating the machine learning paradigm with the Knowledge base system in the healthcare field?

**RQ4:** What are the Challenges & limitations of integrating ML and Symbolic reasoning?

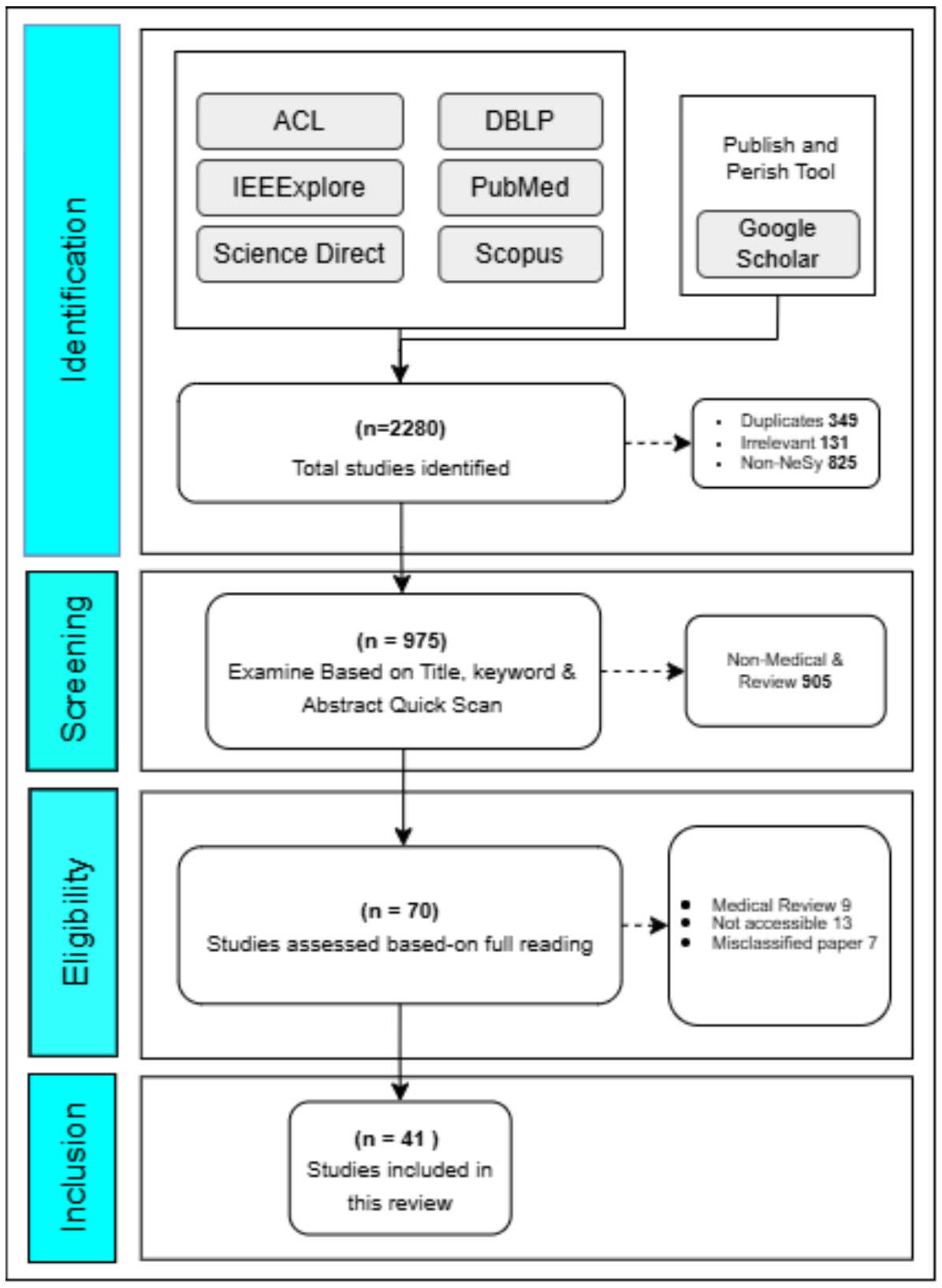

*Fig. 1 NeSy Literature Collection Flow Chart*

To identify relevant studies, an initial search retrieved 2,280 scientific articles from digital databases such as ACL, IEEE Xplore, ScienceDirect, DBLP, PubMed, Scopus, and NeurIPS, ICML Fig.1. For databases like ACL, PubMed, and Google Scholar, we collected by manual search (CSV). DBLP was Jason format and Science Direct BibTex format. We used Paperpile, an online tool mainly used for Science Direct BibTex to CSV conversion. Afterward, we used Python & pandas to handle Jason format and data preprocessing.

In the identification stage, we eliminated duplicates, irrelevant articles, inaccessible sources, and non-medical studies, narrowing down to 975 manuscripts on neuro-symbolic themes. Further screening excluded 905 non-medical articles based on title, keywords, and abstracts. Applying eligibility criteria and research questions, we finalized 41 biomedical-relevant studies. We prioritized the biomedical domain and included promising models and relevant articles potentially applicable to healthcare. From the eligible articles, we extracted details such as (a) Title, (b) Model Performance metrics, f(c) Results, (d) Source/Venue, (e) Domain, (f) Publication Year, (g) Model Reference, (h) Dataset, (i) NeSy category, (j) Integration type, and additional symbolic and sub-symbolic terms.

*Resources & Supplementary materials:*

All statistics, facts, figures, tools, and supporting materials leveraged in this study can be acquired upon requesting to upon request to mhossai5@uab.edu.

## III. NEURO SYMBOLIC ARTIFICIAL INTELLIGENCE

The roots of Neuro-symbolic AI, merging symbolic reasoning and neural processing, can be traced back to McCulloch and Pitts' 1943 innovation [44]. McCulloch and Pitts' work was an avant-garde in its use of propositional logic to model with neural computation. Since the late 1980s, there has been ongoing debate [45] about the need for a cognitive sub-symbolic level in AI. In artificial Intelligence (AI), perception and cognition are represented by two foundational paradigms: connectionism (neural systems) and symbolism, or knowledge-based systems. Each approach has been a dominant influence in the field for decades after decades. Connectionism focuses on pattern recognition and learning but is unable to interpretability and reasoning, and symbolism emphasizes logic and rule-based reasoning but has flaws in learning from big and noisy data. Neuro-symbolic AI combines symbolic reasoning (Knowledge-based, logical systems) with deep learning (data-driven, statistical methods) to create AI systems capable of accomplishing cognitive tasks. Literature evident [27] that the primary goals of Neuro-Symbolic AI (NeSy) include but are not limited to generalization, interpretability, explainability, reasoning, error handling/root cause identification, out-of-distribution handling, scalability with minimal data, transparency, trustworthiness, reduced computational complexity, and energy efficiency. However, an underlying comparison will make it easy to understand both domains.

### A. *Fundamental Comparison*

The fundamental distinctions between symbolic, sub-symbolic, and neuro-symbolic AI reflect their attributes and approaches toward intelligence jobs. Literature suggests [3] Symbolic AI, or symbolism, relies on explicit signs, logical reasoning, serial processing, and static structures respectively. It requires human intervention and precise inputs and focuses on concept composition and model abstraction. By contrast, sub-symbolic AI, represented by connectionism, employs statistical methods and associative, parallel processing in dynamic systems. It's flexible, adapts to noisy or incomplete data, generalizes from large datasets, and supports concept creation through learning. Neuro-symbolic AI aims to merge these paradigms, leveraging the strengths of symbolism.

Reasoning and sub-symbolic learning to achieve attributes like interpretability, scalability, error handling, and generalization, advancing model transparency and robustness with less reliance on extensive data. However, Table 1 illustrates a core comparison among Symbolic, Sub-symbolic, and Neuro-symbolic AI.

**Table 1:** A Comparison Presentation of Symbolic, Sub-symbolic, and Neuro-Symbolic [46]

| **Aspects** | **Symbolic** | **Sub-symbolic** | **Neuro-symbolic** |
|---|---|---|---|
| Methods | Mostly logical and algebraic | Mostly analytical & numeric | A combination of logic and numeric |
| Strengths | Productive, recursion principle, compositionality | Robustness, Learning Ability, Adaptivity | High efficiency, data accuracy, learning & reasoning |
| Weaknesses | Consistency constraints, lower cognitive abilities | Opaqueness, Higher Cognitive Ability | Complex to set up rules with connectionist |
| Applications | Reasoning, problem-solving | Learning, control, vision | Control the quality, detection, classification |
| Cog. Science | Not biologically inspired | Biologically inspired | Integration of genetic-inspired ANN & rules |

## B. *Categories and Integration Strategies*

In the AAAI 2020 Robert S. Engelmore Memorial Award Lecture, Henry Kautz outlined five categories of Neuro-symbolic AI systems to discuss the future of AI [21]. Type 1 (symbolic-neuro-symbolic), Type 2 (symbolic [neuro]), Type 3 (neuro; symbolic), and Types 4 and 5 (neuro: symbolic → neuro and neuro_symbolic) (Table 2). Similarly, W. Wang et al. proposed six neuro-symbolic types [47], while K. Hamilton [27] consolidated these into three: ensemble, sequential, and integrated, merging Kautz's categories.

Table 2: NeSy Kautz's category

| **Kautz's Category** | **Description** | **Reference** |
|---|---|---|
| 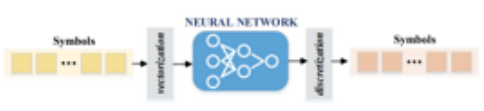  | 1. *Symbolic Neuro Symbolic* | word2vec [177]<br>Glove [178]<br>GPT-3 [179] |
| 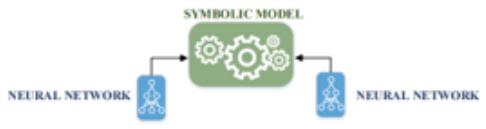  | 2. *Symbolic [Neuro]* | AlphaGo [180]<br>NeSS [181]<br>PLANS [182]<br>VisProg [154]<br>HuggingGPT [183]<br>ViperGPT [184] |
| 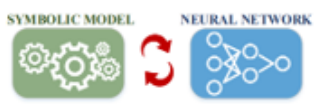  | 3. *Neuro\|Symbolic* | NVSA [185]<br>NeuPSL [186]<br>NSCL [160]<br>NeurASP [20]<br>ABL [187]<br>NSVQA [8] |
| 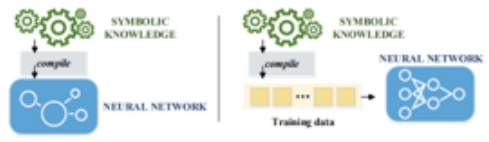  | 4. *Neuro: Symbolic → Neuro* | LNN [10]<br>Symbolic Mathematics [188]<br>Differentiable ILP [189] |
| 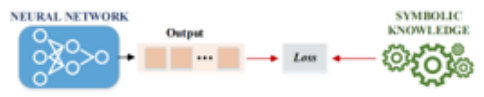  | 5. *$Neuro_{SYMBOLIC}$* | LTN [9]<br>Deep ontology networks [58]<br>Logistic Circuits [190]<br>LRI [191]<br>HDNNLR [192]<br>LFS [193]<br>Semantic Loss [ 61]<br>DANN [194] |
| 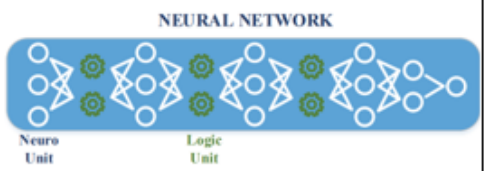  | 6. *Neuro[Symbolic]* | GNN+attention [20]<br>NLM [54]<br>SATNet [195]<br>GVR [196]<br>LogicSeg [197] |

Note: Visual diagram adopted from [39]

Furthermore, the study was published in 2005 by M. Hilario [48]. Neuro-symbolic AI approaches were classified across eight dimensions, grouped into three main aspects: Integrated vs. Hybrid, Neuronal vs. Connectionist, and Local vs. Distributed types. Notably, no single approach was deemed universally superior; rather, the choice of approach depends on specific domain requirements, tasks, and contextual considerations.

Likewise, from a general point of view, two main engines are typically employed in neuro-symbolic integration: symbolic and neural components. Studies underscore knowledge graphs, decision trees, first-order logic [9], fuzzy logic [9], programming languages like Prolog [49], and symbolic expressions [7],[8] as central to the symbolic side. In addition, transformers and large language models (LLMs) are being integrated into the neural component to enhance the system's learning and reasoning capabilities.

Furthermore, the applications of neuro-symbolic AI span diverse domains. IBM Research's "Survey on Applications of Neuro-Symbolic Artificial Intelligence" highlights a range of fields. In this study, we broaden the focus by including models applied in healthcare (the primary focus), as well as cybersecurity, recommendation systems, robotics/automation, smart cities, information retrieval, NLP, VQA, computer vision, digital twins, generative AI, military applications, and marine vessel management. An outline table of application domains, models, and references is available in Appendix A. In addition, Table 3. It serves as a benchmarking summary and performance result for several leading approaches.

## C. *Neuro-Symbolic Reasoning & Explainability*

From a symbolic perspective, reasoning is conducted explicitly through rules or logic, representing the cognitive process of concluding, making decisions, or solving problems based on available information, knowledge, or evidence. Key types of reasoning include deductive, inductive, abductive, analogical, probabilistic, and causal reasoning. In the context of neuro-symbolic AI, Logic Tensor Networks (LTN) [9] integrate logic-based rules expressed in first-order logic (FOL). For example, in healthcare, a rule might state, "If a drug has a Toxicity concentration score below 5, it is safe." Given a drug with a toxicity score of 3, the system concludes that the drug is safe based on this biomedical rule. Knowledge is represented through FOL and integrated via a loss function in LTN, which enhances the system's significance in healthcare by incorporating medical knowledge, reducing the likelihood of misdiagnosis, and enabling healthcare professionals to comprehend the underlying logic. Table 4 below illustrates how different approaches formalize knowledge/rules. Additionally, Explainability aims to make AI decisions and processes understandable to humans, clarifying the reasoning behind a model's predictions or decisions. Several ways can be achieved in explainabilities. For instance, Post-hoc Explanation methods, Feature-level explainability, Rule/Knowledge-Based Explainability, and Process/System Explainability are notable. Explainability in neuro-symbolic AI often relies on knowledge-based components (Symbolic Components). NS-VQA, explainDR, and Greybox-XAI are examples of the NeSy explainable model. Explainability is pivotal in healthcare for building trust and accountability of AI systems.

**Table 3: Benchmarking Summary of Leading Approaches**

| Model | Dataset | Metric | Result | Source/Institution | Business Domain |
|---|---|---|---|---|---|
| NS-VQA [8] | CLEVR | Accuracy | 99.80 | IBM | VQA |
| NS-CL [50] | | Accuracy | 96.90 | DeepMind & MIT | |
| XNMs [57] | | Accuracy | 100.00 | IEEE | |
| Prob-NMN [59] | | Accuracy | 95.63 | MLR Press | |
| DL2 [60] | MNIST | Accuracy | 97.60 | MLR Press | Classification |
| Semantic Loss [61] | | Accuracy | 99.36 | MLR Press | |
| LTN [9] | | Accuracy | 98.04 | Sony AI | |
| Faster LTN [62] | PASCAL VOC | mAP | 73.80 | Springer | Object Detection |
| Neural LP [63] | UMLS, Kinship | MRR | 94.00 | Neurips | Link Prediction |
| RRN [58] | Family Trees | Accuracy | 99.90 | IJCAI | |
| Grail [64] | NELL-995 | AUC-PR | 98.11 | ICML | Relation Prediction |

**Table 4: NeSy Symbolic Expression**

| RULES | Expression | Model |
|---|---|---|
| First-Order Logic | $\forall x_A, p(x_A, l_A), \forall x_B, p(x_B, l_B)$ | LTN |
| Fuzzy Logic | $F = \forall x(\text{isCarnivor}(s)) \rightarrow (\text{is Mammal}(x))$<br>$\{\text{isCarnivor}(s): [0,1], \text{is Mammal}(x): [1,0]\} \rightarrow F = [1,0]$ | LTN |
| Logic Rules | Domain: $animal(dog).carnivore(dog).mammal(dog)$ Logical formula: $mammal(x) \wedge carnivore(x)$<br>ABL: $hypos(x) \coloneqq animal(x), mammal(x), carnivore(x)$ | ABL |
| Problog | $flip(coin1).flip(coin2).$<br>$nn(mside, C, [heads, tails])::$<br>$side(C, heads); side(C, tails).t(0.5):: red; t(0.5):: blue.$<br>$heads :- flip(X), side(X, heads).$<br>$win :- heads.$<br>$win :- + heads, red.$<br>$query(win).$ | DeepProblog |
| Symbolic Expression | $equal_shape: (entry, entry \rightarrow Boolean$<br>$equal_size\ (entry, entry) \rightarrow Boolean$ | NS-VQA |

**Table 5: Selected Models Goals and Principles Summary**

| | Model | Authors | Year | Knowledge Representation | Interpretability | Reasoning | Explainability |
|---|---|---|---|---|---|---|---|
| 1. | **KBANN** | GG Towell et al., | 1994 | Propositional Logic | | ✓ | |
| 2. | **CLIP** | A d'Avila Garcez et al., | 1999 | Propositional Logic | | ✓ | |
| 3. | **NS-VQA** | K Y et al., | 2018 | Symbolic Expression | | | ✓ |
| 4. | **DeepProbLog** | R Manhaeve et al., | 2018 | First-Order Logic | | ✓ | |
| 5. | **NS-CL** | J Mao et al., | 2019 | Symbolic Expression | | ✓ | |
| 6. | **XNMs** | J Sh et al., | 2019 | Knowledge Graph | | | ✓ |
| 7. | **XAI** | M Drance et al., | 2021 | Symbolic Expression | | | ✓ |
| 8. | **ExplainDR** | S I Jang et al., | 2021 | Symbolic Expression | ✓ | | ✓ |
| 9. | **NS-QAPT** | C Ashcraft et al., | 2023 | Symbolic Expression | | | ✓ |
| 10. | **Greybox XAI** | A Bennetot et al., | 2022 | Symbolic Expression | | | ✓ |
| 11. | **X-NeSyL** | N D Rodríguez et al., | 2022 | Symbolic Expression | | | ✓ |
| 12. | **NS-ICF** | W Zhang et al., | 2022 | Symbolic Expression | | | ✓ |
| 13. | **SenticNet** | E Cambria et al., | 2022 | Symbolic Expression | ✓ | | ✓ |
| 14. | **HRI** | C Glanois et al., | 2022 | First-Order Logic | ✓ | | |
| 15. | **LTN** | D Onchis et al., | 2022 | First-Order Logic | | ✓ | |
| 16. | **EIC** | S Haji et al., | 2023 | Symbolic Expression | ✓ | ✓ | |
| 17. | **KD-LTN** | H D Gupta et al., | 2023 | First-Order Logic | ✓ | ✓ | |
| 18. | **GNN** | Raj K et al., | 2023 | Symbolic Expression | ✓ | | |
| 19. | **NEUROSIM** | Singh H et al., | 2023 | Symbolic Expression | ✓ | ✓ | |
| 20. | **NSIL** | Cunnington D et al., | 2023 | Symbolic Expression | ✓ | ✓ | |
| 21. | **TACRED** | R Vacareanu et al., | 2023 | Symbolic Expression | ✓ | ✓ | ✓ |
| 22. | **NSHD** | H Lee et al., | 2023 | Symbolic Expression | | | ✓ |
| 23. | **TON-ViT** | Y Zhuo et al., | 2023 | Symbolic Expression | | | ✓ |
| 24. | **DeepPSL** | S Dasaratha et al., | 2023 | First-Order Logic | | ✓ | ✓ |
| 25. | **S-REINFORCE** | Dutta R et al., | 2024 | Symbolic Expression | ✓ | | |
| 26. | **DVP** | Cheng Han et al., | 2024 | Symbolic Expression | | | ✓ |
| 27. | **X-VQA** | A Mishra et al., | 2024 | Symbolic Expression | | | ✓ |
| 28. | **MARL** | C Subramanian et al., | 2024 | First-Order Logic | ✓ | | |
| 29. | **NS-RL** | L Luo et al., | 2024 | First-Order Logic | ✓ | | ✓ |

Note: A list of existing study's principles, such as reasoning and explainability, was summarized based on reading related articles.

## IV. NEURO SYMBOLIC AI IN HEALTHCARE

Artificial Intelligence in healthcare has transformed clinical domains by enhancing disease surveillance, drug development, diagnostics, prognostics, treatments, cancer research, genomic study, and protein research. A recent breakthrough is the invention of AlphaFold. This revolutionary AI system predicts the 3D proteins' highly accurate structure based on amino acid sequences developed by John Jumper (DeepMind). He was awarded the 2024 Nobel Prize in Chemistry for their work [65]. Despite these promising advances, several challenges remain, including the need for transparent, explainable, robust AI models. The World Health Organization's 2021 guidance on the ethics and governance of AI in healthcare outlines six core principles to ensure AI benefits people globally [66]. The principles of transparency, explainability, and intelligibility are focal and highlighted, emphasizing the need for AI systems. As such, Neuro-symbolic AI promises to revolutionize healthcare by combining the strengths of neural networks (robust pattern recognition) and symbolism (reasoning and explainability). The explainable nature of symbolic methods enhances decision-making by providing clinicians with understandable justifications for AI-driven diagnoses and treatment plans, fostering trust and collaboration. This improves patient care through more reliable, transparent, and interpretable AI systems.

In this survey, various approaches and domains of application discovered that have emerged in Neuro-symbolic AI in the past ten years, advocating great potential regarding knowledge representation, reasoning, explainability, and interpretability. This section aims to present 41 eligible articles on Neuro-symbolic healthcare and their empirical results, with the core findings focusing on their application in the medical domain. Table 6 provides insights into significant models, including their performance metrics, neural and symbolic integration terms, year of publication, and dataset. Furthermore, the following list highlights diverse use cases that demonstrate NeSy's applicability across various medical and biological domains that our study revealed:

Phenotype-driven variant prioritization [67], ontology-based chemical classification[68], LUS-based COVID-19 severity assessment[69], high-order relationships for hyperlink prediction[70], molecular generation[71], mental disorder diagnosis[72], toxicity prediction[73], COVID-19 patient stratification[74], seizure detection[75], link prediction[76]-[78], wound healing stage prediction[79], early depression detection[80], detection and segmentation of cerebral aneurysms in two-dimensional digital subtraction angiography (DSA) images[81], ontological classification[82], E. Coli promoter gene sequences (DNA) classification [6][83], genetic sequence classification[84], classify protein secondary structure prediction[12], protein folding predictions[85][86], pulmonary embolism diagnosis [87], predicting cell states from gene expression profiles [88], probability prediction of age from gene expression records of skin tissue [89], classification of cell types and states [28], treatment of specific diseases [91], heart failure classification [94], question and a relevant piece of medical literature (a context) [93], protein function prediction [13], ontology-based classification[94], target gene prediction [14], predicting interactions between proteins [95], diabetic retinopathy classification [15], semantic classification [96], temporal series classification problem [97], neurodegenerative diseases[16], pattern recognition in cardiotocograms [98], aphasia diagnosis [99], i2b2 2008 obesity challenge classification [19], breast tissues classification [100].

Additionally, integrating data and knowledge enables advancements in various biomedical, chemoinformatics, and applications that can accurately diagnose diseases, discover potential drug targets, de novo drug and artificial protein design, medical imaging, Cardiotoxicity, and Ophthalmology. Some of the practical domain applications of NeSy are as follows:

### A. *Drug Discovery and Cheminformatics*

Usually, the drug development process—from target discovery to regulatory approval—typically spans 12 to 15 years and costs approximately $2.8 billion [101]. Moreover, global drug sales are forecasted to reach $1.9 trillion by 2027 [101]. In the Generative AI Era, an opportunity opens to reduce these timeframes and costs significantly. For instance, models like Llamol have demonstrated impressive accuracy, generating novel valid molecules with 97-99% [102]. However, they rely heavily on data-driven approaches. Combining symbolic reasoning with neural networks, Neuro-Symbolic AI models like Logic Tensor Networks (LTN) (Fig. 2) enhance predictive power; in our trials, LTN and DeepProbLog achieved around 97% accuracy in classifying drug efficacy and bioactivity for diabetes and cancer (TNBC) inhibitors bio-activity classification employed ChEMBL dataset [103], and outperforming conventional DNNs and transformer models like RoBERTa. The adaptability of LTN extends to the following applications: drug sensitivity, synergy, toxicity, resistance, response, multi-target drug discovery, molecular optimization, and binding affinity prediction. In addition, authors M. Drance, and T.T. Ashburn have discovered a new use of existing drugs, which is known as drug repurposing using the NeSy approach [76][104].

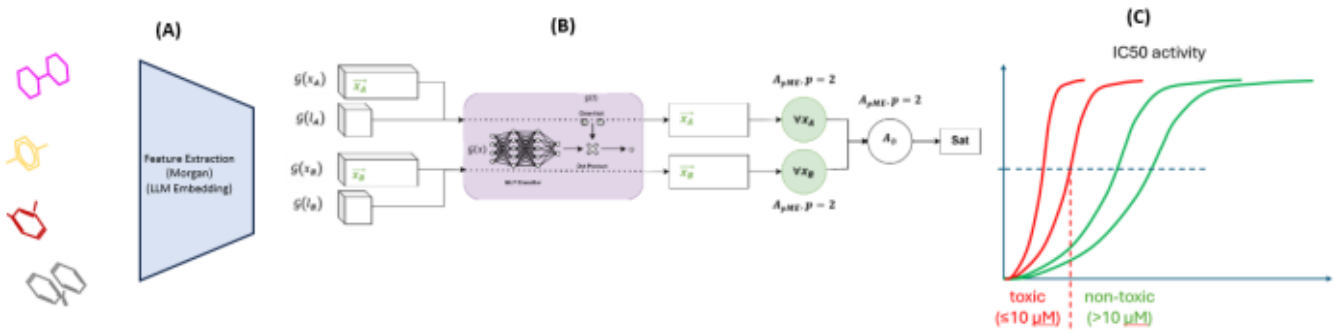

*Fig.2: LTN architecture (B)*

### B. *Protein Research*

Proteins play an essential role in biology, from blocking infections to harnessing solar energy, vision, blood clotting, immune system response, hormone regulation, and cell and tissue repair. Understudying the correct structures and highly accurate folding is vital. Unfolded or misfolded proteins may cause degenerative diseases, including Huntington's diseases, Alzheimer's, Parkinson's, and cystic fibrosis. Recent AlphaFold's groundbreaking innovations have ushered in a new era of computational protein engineering research, delivering highly accurate protein structure predictions. Prof. David Baker from the University of Washington received a Nobel Prize for pioneering de novo protein design [65], an accomplishment that allows scientists to create custom proteins from scratch. In the realm of Neuro-Symbolic AI, research can be expanded into protein folding prediction [85][86], protein structure prediction[12], and protein function analysis. By integrating AlphaFold's capabilities with symbolic reasoning, researchers can enhance the precision of protein-protein interaction predictions that can unlock new insights for targeted drug design and synthetic biology applications.

### C. *Visual Question Answering (VQA)*

Transparency and interpretability are keys in healthcare that improve reliability and trustworthiness in the AI system. Integrating medical domain-specific knowledge with neural networks [105][106] is an emerging field that enables more accurate, context-aware predictions and enhances the interpretability of AI models, flooring the way for improved diagnostics, personalized treatments, and advanced therapeutic discoveries. NeSy approach has made significant contributions in this domain, identifying more than 51 robust, visual reasoning, interpretable, and explainable models [107]-[157]. To be exemplified, NS-VQA [8], developed by IBM, MIT, and DeepMind, combines symbolic reasoning with deep learning to answer questions about images. It interprets visual scenes by translating them into symbolic representations, allowing for logical reasoning over objects, and achieved 99.80% accuracy on CLEVR benchmarks, whereas the NS-CL model yields 96.90% accuracy [50]. In this field, some frequently used benchmarks datasets are VQA 1.0 dataset [161], CLEVR datasets [159], GQA dataset [162], CLEVR-CoGenT [159], and SHAPES dataset [163], selected models that can be translatable into Med-VQA such as NS-VQA, XNMs[57], NMN[56], N2NMN[164], NSM[165], VRP [176].

### D. *Ophthalmology & Cardiotoxicity*

In ophthalmology, Neuro-symbolic AI combines symbolic reasoning with pattern recognition to enhance diagnostic and treatment explainability, enabling interpretable and accurate assessments of eye conditions such as diabetic retinopathy and glaucoma. This approach improves detection accuracy and adds transparency, aiding ophthalmologists in understanding AI-driven decisions. ExplainDR[15] uses a feature vector as a symbolic representation for explainable diabetic retinopathy (DR) classification Fig. 3. Additionally, the NeSyL[34] study proposes viable neuro-symbolic approaches for clinical settings with potential applications in diagnosing and predicting ocular diseases and in health informatics.

Furthermore, we conducted simulations assessing the cardiotoxic effects of drug molecules using Logic Tensor Networks (LTN) Fig. 3. We constructed a hERG-related dataset by combining ChEMBL[103], hERG Karim[167], BindingDB [168], PubChem[[169], hERG Blockers[110], and GTP datasets[171]. LTN outperformed models such as M-PNN [172], OCHEM Predictor-II [167], Random Forest [173], CardioTox[167], ADMETlab 2.0[174], and Gradient Boosting[175], achieving higher accuracy, with an ACC of 0.827 and a specificity (SPE) of 0.890 on the hERG-70 benchmark [202] (Fig., 2).

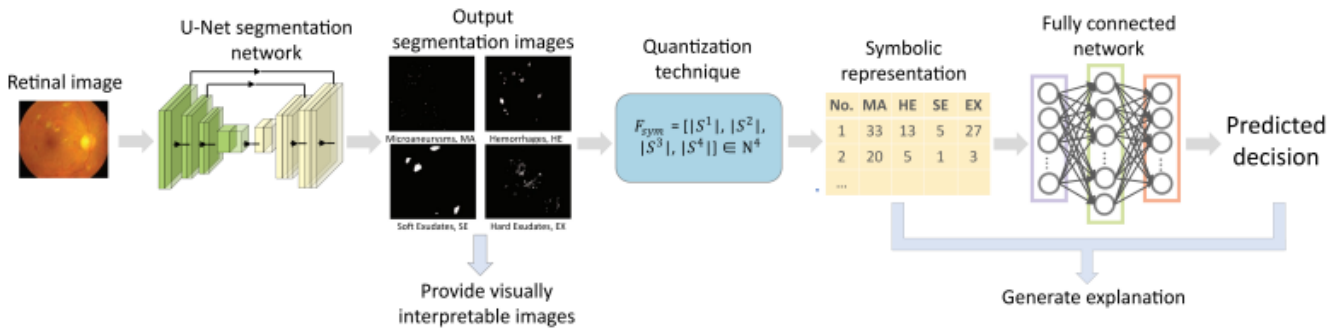

*Fig. 3: ExplainDR Architecture*

**Table 6:** An overview of recent NeSy models in healthcare with performance metrics

| | Method | Model Inte | Dataset | Metrics Type | Result | Year | Venue/Source | Reference |
|---|---|---|---|---|---|---|---|---|
| 1. | KBANN | PR+DL | Promoter Dataset | Error Rate | 4/106 | 1990 | ACM | [83] |
| 2. | KBANN | PR+DL | Splice-Junctions | Error Rate | 7.50% | 1992 | ScienceDirect | [6] |
| 3. | KBANN | PR+DL | splice-junctions [DNA] | Error Rate | 6.4% | 1990 | Neurips | [84] |
| 4. | FSKBANN | PR+ non-Recursive +DL | Data set from Qian and Sejnowski (1988) | Accuracy | 63.4% | 1992 | ACM | [12] |
| 5. | FSKBANN | PR+ Non-RL+DL | The data set consists of 128 proteins. (85/43) Tarin/Test | Accuracy | 61.9% | 1991 | ScienceDirect | [85] |
| 6. | KBANN | PR +DL | 16 in vivo 31P MR spectra | Average Pat / Error | 0.1179 | 1998 | Citeseerx | [86] |
| 7. | KBANN | DL/MLP | PE diagnosis dataset | Accuracy | 100% | 2007 | PubMed | [87] |
| 8. | KPNN | DL+UI | TCR dataset | ROC AUC | 98.4% (interquartil | 2020 | PubMed | [88] |
| | | | | | e range, 0.979 to 0.987) | | | |
| 9. | KPANN | DL+UI | Transcriptomic dataset | MAE, Median abs. error | 5.51 y, 4.71 y | 2021 | Nature | [89] |
| 10. | Pathway-Primed | DL+PK | Mouse dataset, Human dataset | Accuracy | 0.936 | 2022 | IDUS | [90] |
| 11. | LTN | DL+FOL+FL | Mixed [Neoplasm, Glaucoma] | Accuracy | 81-80 | 2021 | IJCLR | [91] |
| 12. | BaLONN | DL+UI | Heart Failure Clinical Records | Tot. PR (%) | 69.28% | 2021 | Europe PMC | [92] |
| 13. | BioBERT+MDAtt | DL+MDAtt | BioBERT QA, PubMed QA | Accuracy | 84% | 2021 | Oxford Academic | [93] |
| 14. | MultiPredGO | DL+UI | Protein Sequence | AUC | 0.8169 | 2021 | IEEE Xplore | [13] |
| 15. | N/A | ML+SPARQL | GO | ROCAUC | 0.79 | 2017 | Oxford Academic | [94] |
| 16. | DeepMiR2GO | DL+UI | miRNA2GO-337 | F1-Max | 39.90% | 2019 | MDPI | [14] |
| 17. | SiameseNN (ont) | ML+FOL | PPI Dataset | AUC | 0.91% | 2021 | Oxford Academic | [95] |
| 18. | ExplainDR | DL | IDRiD Dataset | Accuracy | 60.19% | 2021 | CEUR-WS | [15] |
| 19. | RLR | DRL+DR | BIKG Hetionet | MRR | 0.167±.008 | 2021 | arXiv | [77] |
| 20. | Ne-Sy | DL+UI | Ontologies Dataset | has-indication[{F-Measure}] | 72% | 2018 | CEUR-WS | [78] |
| 21. | Ne-Sy | DL+UI | COVID-19 LUS dataset | MPA | 82% | 2022 | Harvard Press | [74] |
| 22. | N/A | ML+NR | EEG dataset | Accuracy [AVG]-β | 68% | 2022 | ACM | [97] |
| 23. | PP-DKL | DL+PPL | ADNI | MAE[MMSE] | 1.36±0.27 | 2021 | Springer | [16] |
| 24. | PoLo | DL+ Logic rules | OREGANO KG | MRR | 0.498 | 2021 | Scitepress | [76] |
| 25. | FSD | DL+ Fuzzy Logic | N/A | None | None | 2008 | Springer | [17] |
| 26. | Neuro-Fuzzy Approaches | Pattern Recognition + UI | FHR Signal | Accelerations Accuracy<br>Decelerations Accuracy | 72.60%<br>65.70% | 2002 | Springer | [98] |
| 27. | N/A | ML+ Certainty Factor Rules | Aphasia Diagnosis | F-Measure | 80% | 2019 | IEEE | [99] |
| 28. | KGCNN | DL+UI | I2B2-VA challenge dataset | Macro F1 (Intuitive)<br>Micro F1(Intuitive) | 0.6760<br>0.9613 | 2018 | IEEE | [19] |
| 29. | KBANN | DL+PR | vivo 31P MR spectra | ACG Pat/Error (cKBANN-5) | 0.1248 | 1999 | IEEE | [100] |
| 30. | EmbedPVP | KG+DL | PAVS database | ROCAUC | 0.9960 | 2024 | Oxford Academic | [67] |
| 31. | ChEB-AI | LNN+DL | ChEBI | F1 (micro) | 0.9032 | 2024 | RSC | [68] |
| 32. | N/A | DT+DNN | Covid-19 LUS Dataset | Prognostic value (max) | 95.00 | 2023 | Elsevier | [69] |
| 33. | NeSyKHG | KHG+DL | Chinese Medical Highorder Relational (CMHR) dataset | Accuracy | 0.947 | 2024 | Elsevier | [70] |
| 34. | S-Reinforcement | Subsymbolic+DL | SMILES Data | Reward | 9.50 | 2024 | IEEE Xplore | [71] |
| 35. | LNN | Subsymbolic+DL | Mental disorder diagnosis | AUC | 0.76 | 2023 | ACL | [72] |
| 36. | ChEBai | Ontology+DL | ChEBI, Tox21 | F1 (micro) | 0.86 | 2023 | Springer | [73] |
| 37. | STONE | Temporal Logic+DL | CHB-MIT database | F1 | 0.9834 | 2022 | Elsevier | [75] |
| 38. | NSTSC | Temporal Logic+DL | UCR TIME-SERIES DATA ARCHIVE | Accuracy | 84.66 | 2022 | IEEE | [79] |
| 39. | TAM-SenticNet | SenticNet+DL | CLEF eRisk 2022 Lab | F1 score | 0.758 | 2024 | Elsevier | [80] |
| 40. | DeepInfusion | Knowledge Extraction + DL | intracranial aneurysms in-house prepared dataset of 409 DSA images | IOU | 0.9602 | 2023 | Elsevier | [81] |
| 41. | ELECTRA | Ontology+DL | ChEBI Dataset | F1 Score | 0.78 | 2023 | IOS Press | [82] |

*Note*: PR - Propositional Rule; Non-RL - Non-Recursive Rule; UI - Un-identified; MAE - Mean Absolute Error; DR - Deterministic Rules; MRR - Mean Reciprocal Rank; MPA - Mean Prognostic Agreement; NR - Neuroaesthetics Rules; PPL - Probabilistic Programming Languages; ADNI - Alzheimer's Disease Neuroimaging Initiative; Approach - Author does not mention method or model name, thus labeled as an approach; DL - Deep Learning; RL - Reinforcement Learning; FOL - First Order Logic; FL - Fuzzy Logic; ML - Machine Learning; PPI - Protein-Protein Interaction; DRL - Deep Reinforcement Learning; DR - Drug Repurposing; KG - Knowledge Graph; LNN - Logic Neural Network; DT - Decision Tree; DNN - Deep Neural Network; LUS - Lung Ultrasound; AUC - Area Under the Curve; ROC - Receiver Operating Characteristic; ACC - Accuracy.

## V. NEURO-SYMBOLIC CHALLENGES

Despite the promise demonstrated by neuro-symbolic (NeSy) approaches in various applications, several critical challenges persist in advancing their use in biomedical contexts, as outlined below.

*Knowledge Representation*: Encoding biomedical knowledge, such as protein-protein interactions for disease diagnosis, in NeSy models is challenging due to the depth and complexity of information needed for precise clinical applications.

*Lack of Standardized Benchmarks & Evaluation*: The absence of consistent benchmarks for NeSy systems, like those for drug interaction prediction, hinders the assessment of model efficacy across different healthcare scenarios.

*Handling Adversarial Attacks*: NeSy models in healthcare, such as those used for diagnostic image analysis, must be resilient against adversarial attacks, which could exploit model vulnerabilities and potentially lead to misdiagnoses.

*Common Sense Reasoning*: Integrating common sense reasoning, like understanding disease progression stages, is essential for NeSy models to make accurate inferences in complex cases, such as cancer prognosis.KG can be a good candidate to develop robust symbolic engineering.

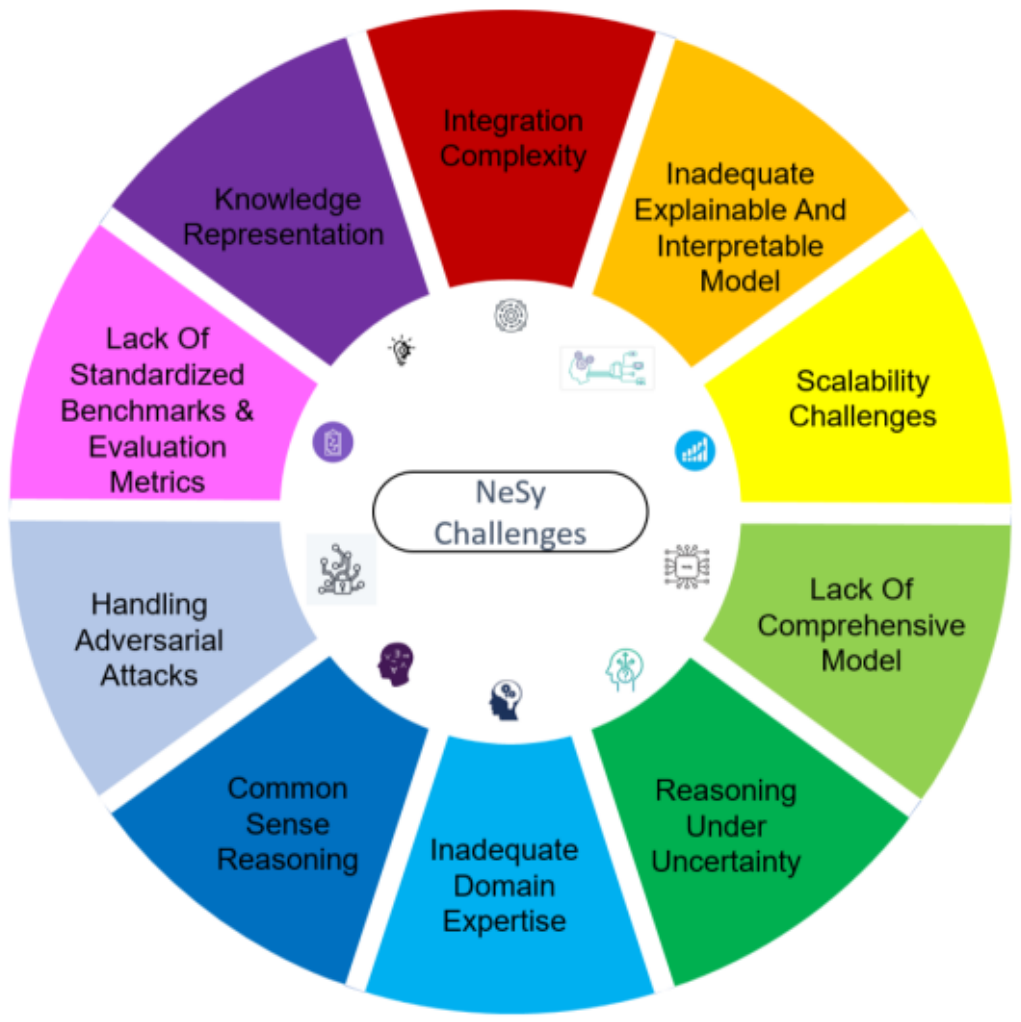

*Fig. 4: Neuro-Symbolic Open Challenges*

*Integration Complexity*: Combining symbolic reasoning and neural networks for medical imaging, like X-ray analysis with rule-based interpretations, is complex and affects model reliability scalability for diverse clinical settings.

*Inadequate Explainable and Interpretable Models*: In bioactivity prediction for drug development, NeSy models need to explain why certain compounds are predicted to interact with biological targets. Additionally, inadequate *Explainable notable.*

*Scalability Challenges*: Scaling NeSy models to handle large genomic datasets, such as genome-wide association studies (GWAS) for disease traits, is difficult since there is a trade-off between accuracy and explainability while maintaining large-scale prediction.

*Lack of Comprehensive Model*: Existing NeSy models are often limited in scope(lack of regression model) and may struggle to handle the diverse range of tasks needed in healthcare, such as simultaneously predicting drug efficacy and toxicity.

*Reasoning Under Uncertainty*: In clinical decision support systems, NeSy models must make predictions with uncertain or incomplete patient data, such as when predicting outcomes for rare diseases.

*Domain Expertise*: Building NeSy models for fields like personalized medicine requires high domain expertise, especially when interpreting complex genomic and phenotypic data.

## VI. FUTURE DIRECTION AND OPEN PROBLEM

We propose future research directions building upon the current findings to address existing limitations and further enhance the potential of neuro-symbolic models in biomedical applications, such as:

### A. *Proposed Novel Approach (LTN-CPI) Using Chemical, Protein Language, and LTN Models for Compound-Protein Interaction (CPI) prediction:*

Protein-based therapeutics have emerged as one of the most rapidly advancing sectors within the pharmaceutical industry, driving transformative paradigms in disease treatment. There was an anticipated to account for half of the top ten best-selling drugs globally [198] based on targeting protein. Classical therapeutic drug targets are predominantly concentrated within approximately 130 protein families [199], and the therapeutic protein market has surpassed a valuation of USD 380 billion [200]. Given their significance, studying drug/compound and protein interactions is critical. Research in Compound-Protein Interactions (CPI) or Drug-Target Interactions (DTI) has been mainly instrumental, contributing to substantial advancements in this field [201]. We propose a new reasoning-capable approach for CPI use cases that can be accomplished in breast cancer or any inhibitor- or no-inhibitor-based CPI prediction task.

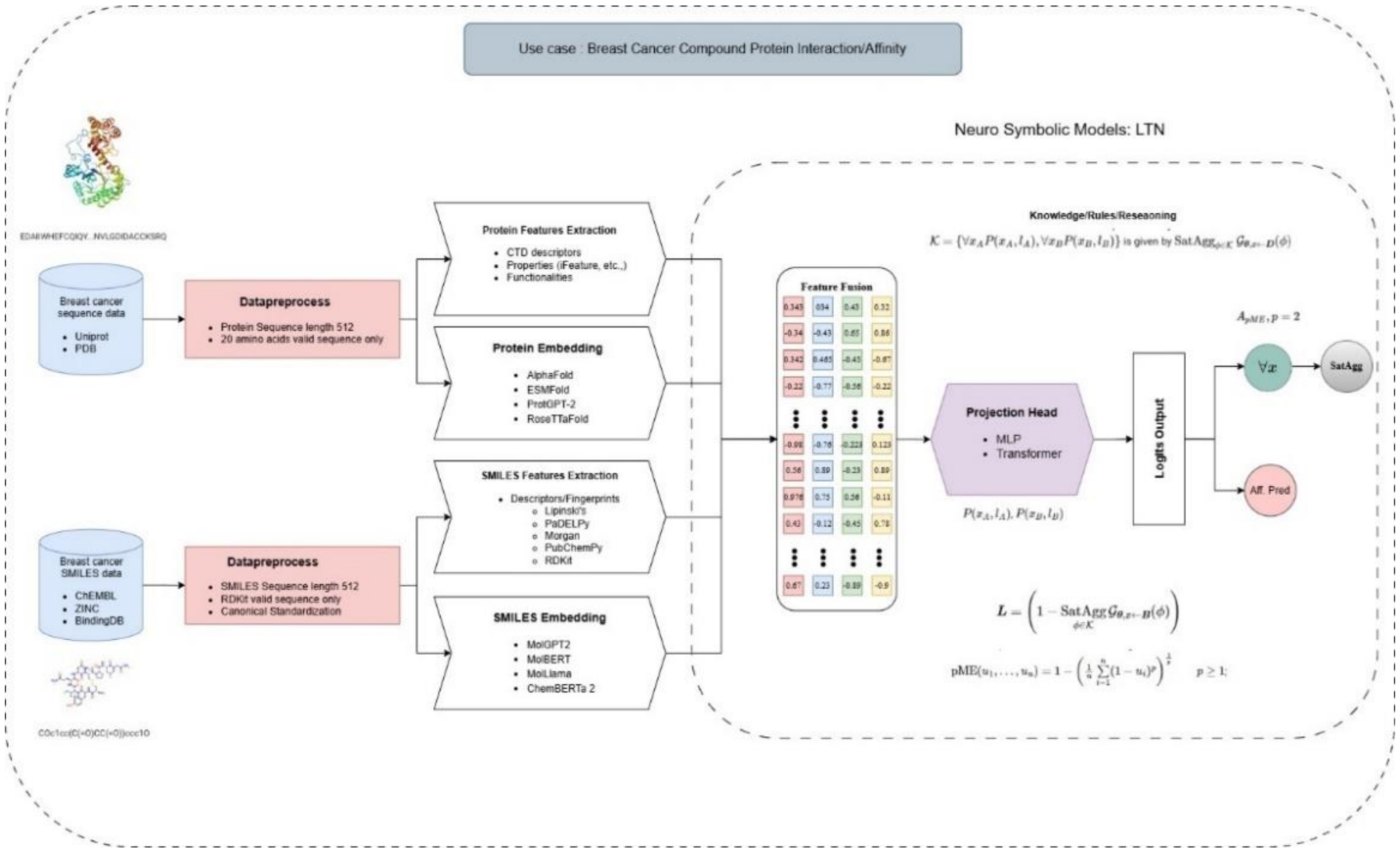

*Fig. 5: Approach Proposed (LTN-CPI) Using Chemical, Protein Language, and LTN for Compound-Protein Interaction (CPI) Prediction*

Table 7 illustrates the rules, learning, and loss according to LTN constructions in the CPI/DTI task context. The proposed approaches can be experimented with utilizing BindingDB, DAVIS, and KIBA benchmarks from the TDC Drug Target Interaction (DTI) porch.

**Table 7:** LTN Knowledge-based Setting for Compound-Protein Interaction

| **Contents** | **Classification** |
|---|---|
| Define Axioms | • $\forall \mathrm{x}_A, p(\mathrm{x}_A, l_A)$: all the examples of class $A(Active/Postive\ = 0)$should have a label $l_A$ <br> • $\forall x_B, p(x_B, l_B)$: all the examples of class $B$ (Inactive/Negative = 1) should have a label $l_B$ |
| Axioms (rules, knowledge base) | $\mathcal{K} = \forall x_A p(x_A, l_A), \forall x_B p(x_B, l_B)$ |
| SatAgg is given by | $\mathrm{SatAgg}_{\phi \in \mathcal{K}} \mathcal{G}_{\boldsymbol{\theta}, x \leftarrow \boldsymbol{D}}(\phi)$ |
| Learning & Loss | $\boldsymbol{L} = \left(1 - \underset{\phi \in \mathcal{K}}{\mathrm{SatAgg}} \mathcal{G}_{\boldsymbol{\theta}, x \leftarrow \boldsymbol{B}}(\phi)\right)$ |
| **Note**: This table was developed inspired by the official LTN | |

### B. *Inhibitor Prediction and Bio-Activity Classification Using LTN-Enhanced Transformers:*

Logic Tensor Networks (LTN) combined with transformer models can offer enhanced predictive capabilities for chemical classification and bio-activity prediction. This approach could facilitate inhibitor prediction and enable accurate classification of molecular activity based on structural and functional characteristics, benefiting areas like cancer treatment and antimicrobial drug development. The LTN + Transformer model is up-and-coming for applications requiring precise classification and interpretability, such as identifying inhibitors for specific targets in complex biochemical pathways.

### C. *Developing Explainable Med-VQA for Interpretive Systems in Oncology*

In the visual question answering (VQA) context, combining vision-based LLMs with symbolic reasoning can support the development of interpretable systems for the Med-VQA field, such as reasoning explainable prediction of medical imaging. For example, reasoning-capable systems could improve interpretability in triple-negative breast cancer imaging by integrating symbolic frameworks to analyze visual data alongside patient histories. This approach seeks to create a robust reasoning system that can provide explanatory insights, supporting oncologists in diagnosing and strategizing treatments with higher precision.

## VII. Conclusion

This study explored the transformative potential of Neuro-symbolic AI (NeSy) in the healthcare sector. By integrating the strengths of neural networks with symbolic reasoning, NeSy offers unique attributes, such as enhanced explainability and reasoning. We extensively reviewed 977 original Neuro-symbolic studies and identified significant advancements and applications in healthcare and revealed practical simulation results in drug discovery. Moreover, evaluated 41 promising use cases and applications within the biomedical domain. Significantly, we proposed a novel reasoning-capable architecture for compound and drug interaction tasks by integrating chemical and protein language models with the Logic Tensor Network framework.

Overall, these findings underscore the growing potential of Neuro-symbolic AI to bridge critical gaps in explainability, interpretability, and reasoning in AI applications. Building upon this foundation, AI scientists envision hybridizing symbolism and connectionism, the two foundational paradigms of AI. Such an approach could enable machines to achieve human-like cognitive behaviors, including commonsense reasoning, fostering more transparent, trustworthy, scalable, and robust AI-driven healthcare systems. This advancement has the potential to revolutionize various aspects of healthcare, from drug discovery to patient care, ultimately benefiting the global healthcare community.

## VIII. Acknowledgment

We sincerely thank the SPARC and Mayo Clinic for their invaluable support and guidance. We are incredibly grateful to the Computer Science Defense Committee (UAB) members for their insightful feedback, and anonymous friends for helping discussion and to format the article. This project was generously funded through Grant OT2:lOT20D032742-Ol and UMI TR004771.

**Author contributions**

DH conceptualized and wrote the manuscripts. Dr. JC guided, review, and lead the project.

**Funding**

There was no external funding besides the above.

**Availability of data and materials**

Procured 977 papers, and 41 healthcare-related articles, tables, and figures can be obtained upon request to mhossai5@uab.edu.

**Declarations**

**Conflict of interest**: The authors declare no conflict of interest.

**Ethics approval**: Not applicable.

**Consent to participate**: Not applicable.

**Consent for publication**: The authors consent to the publication of this work.

## References

[1] R. E. Uhrig, "Introduction to artificial neural networks," IEEE Xplore, 1995
[2] E. Gibney, D. Castelvecchi, "Physics Nobel scooped by machine-learning pioneers," nature, 2024
[3] E. Ilkoua, M. Koutraki, "Symbolic Vs. Sub-symbolic AI Methods: Friends or Enemies?," CEUR-WS, 2020
[4] Y. LeCun, Y. Bengio, G. Hinton, "Deep learning," nature, 2015
[5] H. Taud, J. F. Mas, "Multilayer Perceptron (MLP)," Springer, 2017
[6] G. G. Towell, J. W. Shavlik, "Knowledge-Based Artificial Neural Networks," Elsevier, 1992
[7] G. G. Towell, J. W. Shavlik, "Interpretation of Artificial Neural Networks: Mapping Knowledge-Based Neural Networks into Rules," Citeseer, 1991
[8] K. Yi, J. Wu, C. Gan, et al., "Neural-Symbolic VQA: Disentangling Reasoning from Vision and Language Understanding," ACM, 2018
[9] S. Badreddine, A. d'Avila Garcez, L. Serafini, et al., "Logic Tensor Networks," Elsevier, 2021
[10] R. Riegel, A. Gray, F. Luus, et al., "Logical Neural Networks," arXiv, 2020
[11] S. Amizadeh, H. Palangi, O. Polozov, et al., "Neuro-Symbolic Visual Reasoning: Disentangling "Visual" from "Reasoning"," MLR Press, 2020
[12] R. Maclin, J. W. Shavlik, "Refining Algorithms With Knowledge-Based Neural Networks: Improving The Chou-Fasman Algorithm For Protein Folding," ACM, 1992
[13] S. J. Giri, P. Dutta, P. Halani, et al., "Multipredgo: Deep Multi-Modal Protein Function Prediction By Amalgamating Protein Structure, Sequence, And Interaction Information," IEEE Xplore, 2021
[14] J. Wang, J. Zhang, Y. Cai, et al., "Deepmir2Go: Inferring Functions of Human Micrornas Using A Deep Multi-Label Classification Model," MDPI, 2019
[15] S. I. Jang, M. J. A. Girard, A. H. Thiery, "Explainable Diabetic Retinopathy Classification Based on Neural-Symbolic Learning," CEUR-WS, 2021
[16] A. Lavin, "Neuro-Symbolic Neurodegenerative Disease Modeling As Probabilistic Programmed Deep Kernels," Springer, 2021
[17] K. Dobosz, W. Duch, "Fuzzy Symbolic Dynamics for Neurodynamical Systems," Springer, 2008
[18] F. Arabshahi, J. Lee, M. Gawarecki, et al., "Conversational Neuro-Symbolic Commonsense Reasoning," AAAI, 2021
[19] L. Yao, C. Mao, Y. Luo, "Clinical text classification with rule-based features and knowledge-guided convolutional neural networks," Springer, 2018
[20] Z. Yang, A. Ishay, J. Lee, "NeurASP: Embracing Neural Networks into Answer Set Programming," IJCAI, 2023
[21] Md Kamruzzaman Saker, L. Zhou, A. Eberhart, et al., "Neuro-symbolic Artificial Intelligence Current Trends," 2021
[22] T. R. Besold, A. S. A. Garcez, S. Bader, et al., "Neural-Symbolic Learning and Reasoning: A Survey and Interpretation," 2017
[23] D. Yu, Bo Yang, Dayou Liu, et al., "A survey on neural-symbolic systems," Harvard, 2021
[24] S. Srikant, U. May O'Reilly, "Can Cognitive Neuroscience inform Neuro-symbolic Models?" MIT, 2021
[25] A. Lyubovsky, A. Madaan, Y. Yang, "Characterizing Neuro-symbolic Reasoning in NLP," ACL
[26] L. C. Lamb, A. Garcez, M. Gori, et al., "Graph Neural Networks Meet Neural-Symbolic Computing: A Survey and Perspective," IJCAI, 2020
[27] K. Hamilton, A. Nayak, B. Bozic, et al., "Is Neuro-symbolic AI Meeting its Promise in Natural Language Processing? A Structured Review," IOS Press, 2022

[28] A. A. Garcez, M. Gori, L. C. Lamb, et al., "Neural-Symbolic Computing: An Effective Methodology for Principled Integration of Machine Learning and Reasoning," DBLP, 2021
[29] W. Gibaut, L. Pereira, F. Grassiotto, et al., "Neurosymbolic AI and its Taxonomy: a survey," arXiv, 2023
[30] Z. Susskind, B. Arden, L. K. John, et al., "Neuro-symbolic AI: An Emerging Class of AI Workloads and their Characterization," DBLP, 2021
[31] A. Garcez, L. C. Lamb, "Neurosymbolic AI: the 3rd wave," Springer, 2023
[32] P. Hitzler, A. Eberhart, M. Ebrahimi, et al., "Neuro-symbolic approaches in artificial intelligence," Oxford, 2022
[33] A. Oltramari, J. Francis, C. Henson, et al., "Neuro-symbolic Architectures for Context Understanding," IOS Press, 2022
[34] M. Hassan, H. Guan, A. Melliou, et al., "Neuro-symbolic Learning: Principles And Applications In Ophthalmology," DBLP, 2022
[35] J. M. Corchado, M. L. Borrajo, M. A. Pellicer, et al., "Neuro-symbolic System for Business Internal Control," Springer, 2005
[36] I. Berlot-Attwell, "Neuro-symbolic VQA: A review from the perspective of AGI desiderata," arXiv, 2021
[37] D. Bouneffouf, C. C. Aggarwal, "Survey on Applications of Neurosymbolic Artificial Intelligence," arXiv, 2022
[38] V. Belle, "Symbolic Logic Meets Machine Learning: A Brief Survey in Infinite Domains," Springer, 2020
[39] W. Wang, Y. Yang, W. Fei, "Towards Data-and Knowledge-Driven AI: A Survey on Neuro-symbolic Computing," IEEE, 2022
[40] L. D. Raedt, S. Dumancic, R. Manhaeve, et al., "From Statistical Relational to Neuro-symbolic Artificial Intelligence," 2020
[41] P. Hitzler, "Neuro-symbolic Artificial Intelligence: The State of the Art," IOS Press, 2022
[42] A. Sheth, K. Roy, M. Gaur, "Neurosymbolic AI - Why, What, and How," IEEE Xplore, 2023
[43] D. H. Hagos, D. B. Rawat, "Neuro-Symbolic AI for Military Applications,", IEEE Xplore, 2024
[44] W.S. McCulloch, W. Pitts, A logical calculus of the ideas immanent in nervous activity, Springer, 1943
[45] M. Frixione, G. Spinelli, S. Gaglio, "Symbols and subsymbols for representing knowledge: a catalogue raisonne," ACM, 1989
[46] K.U. K¨uhnberger, H. Gust, P. Geibel, "Perspectives of Neuro–Symbolic Integration – Extended Abstract," University of Osnabruck, 2008
[47] W. Wang, Y. Yang, F. Wu, "Towards Data-and Knowledge-Driven AI: A Survey on Neuro-Symbolic Computing," IEEE Xplore, 2024
[48] M. Hilario, "An overview of strategies for Neuro-symbolic integration," 1997
[49] R. Manhaeve, S. Dumancic, A. Kimmig, et al., "DeepProbLog: Neural Probabilistic Logic Programming," NeurlPS, 2018
[50] J. Mao, C. Gan, P. Kohli, et al., "The Neuro-symbolic Concept Learner: Interpreting Scenes, Words, and Sentences From Natural Supervision," DBLP, 2023
[51] A.S. Avila Garcez, G. Zaverucha, "The Connectionist Inductive Learning and Logic Programming System," ACM, 1999
[52] A.S. Avila Garcez, L.C. Lamb, "A connectionist computational model for epistemic and temporal reasoning," MIT Press, 2006
[53] K. Yi, C. Gan, Y. Li et al, "CLEVRER: CoLlision Events for Video REpresentation and Reasoning," In ICLR, 2020
[54] H. Dong, J. Mao, T. Lin et al, "Neural logic machines," arXiv, 2019
[55] R. Socher, D. Chen, C. D. Manning, A. Ng, "Reasoning with neural tensor networks for knowledge base completion," ACM, 2013
[56] J. Andreas, M. Rohrbach, T. Darrell et al., "Neural Module Networks," CVF, 2016
[57] J. Shi, H. Zhang, J. Li, "Explainable and Explicit Visual Reasoning over Scene Graphs," IEEE Xplore, 2019
[58] P. Hohenecker, T. Lukasiewicz, "Ontology Reasoning with Deep Neural Networks," IJCAI, 2020
[59] R. Vedantam, K. Desai, S. Lee, et al., "Probabilistic Neural-symbolic Models for Interpretable Visual Question Answering," MLR Press, 2019
[60] M. Fischer, M. Balunovic, D. D. Cohen, et al., "DL2: Training and Querying Neural Networks with Logic", MLR Press, 2019
[61] J. Xu, Z. Zhang, T. Friedman, et al., "A Semantic Loss Function for Deep Learning with Symbolic Knowledge," MLR Press, 2018
[62] F. Manigrasso, F. D. Miro, L. Morra, et al., "Faster-LTN: A Neuro-symbolic, End-to-End Object Detection Architecture," Springer, 2021
[63] F. Yang, Z. Yang, W. W. Cohen, "Differentiable Learning of Logical Rules for Knowledge Base Reasoning," Neurips, 2017
[64] K. K. Teru, E. Denis, W. L. Hamilton, "Inductive Relation Prediction by Subgraph Reasoning," ICML, 2020
[65] E. Callaway, "Chemistry Nobel goes to developers of AlphaFold AI that predicts protein structures", nature, 2024.
[66] WHO, "Ethics and governance of artificial intelligence for health: WHO guidance," WHO, 2021.
[67] A. Althagafi, F. Zhapa-Camacho, R. Hoehndorf, "Prioritizing genomic variants through neuro-symbolic, knowledge-enhanced learning", Oxford Academic, 2024.
[68] M. Glauer, F. Neuhaus, S. Flügel et al., "Chebifier: automating semantic classification in ChEBI to accelerate data-driven discovery", RSC, 2024.
[69] L. L. Custode, F. Mento, F. Tursi et al., "Multi-objective automatic analysis of lung ultrasound data from COVID-19 patients employing deep learning and decision trees", Elsevier, 2023.
[70] B. P. Bhuyana, T. P. Singh, R. Tomard et al., "NeSyKHG: Neuro-Symbolic Knowledge Hypergraphs", Elsevier, 2024.
[71] R. Dutta, Q. Wang, A. Singh et al., "Interpretable Policy Extraction with Neuro-Symbolic Reinforcement Learning", IEEE Xplore, 2024.
[72] Y. Toleubay, D. J. Agravante, D. Kimura et al., "Utterance Classification with Logical Neural Network: Explainable AI for Mental Disorder Diagnosis", ACL, 2023.
[73] M. Glauer, F. Neuhaus, T. Mossakowski et al., "Ontology Pre-training for Poison Prediction", Springer, 2023.
[74] L. L. Custode, F. Mento, S. Afrakhteh et al., "Neuro-symbolic interpretable AI for automatic COVID-19 patient-stratification based on standardized lung ultrasound data", AIP, 2022.
[75] R. Yan, A. A. Julius, "Interpretable seizure detection with signal temporal logic neural network", Elsevier, 2022.
[76] M. Drancé, M. Boudin, F. Mougin et al., "Neuro-symbolic XAI for Computational Drug Repurposing", ScitePress, 2021.
[77] G. Edwards, S. Nilsson, B. Rozemberczki et al., "Explainable Biomedical Recommendations Via Reinforcement Learning Reasoning On Knowledge Graphs", arXiv, 2021.
[78] A. Agibetov, M. Samwald, "Fast And Scalable Learning Of Neuro-Symbolic Representations Of Biomedical Knowledge", CEUR-WS, 2018.
[79] R. Yan, T. Ma, A. Fokoue et al., "Neuro-symbolic Models for Interpretable Time Series Classification using Temporal Logic Description", IEEE, 2022.
[80] R. Dou, X. Kang, "TAM-SenticNet: A Neuro-Symbolic AI approach for early depression detection via social media analysis", Elsevier, 2024.
[81] I. Abdullah, A. Javed, K. M. Malik et al., "DeepInfusion: A dynamic infusion based-neuro-symbolic AI model for segmentation of intracranial aneurysms", Elsevier, 2023.
[82] M. Glauer, T. Mossakowski, F. Neuhaus et al., "Neuro-symbolic semantic learning for chemistry", IOS Press, 2023.
[83] G. G. Towell, J. W. Shavlik, M. O. Noordewier, "Refinement Of Approximate Domain Theories By Knowledge-Based Neural Networks", ACM, 1990.
[84] M. Noordewier, G. Towell, J. Shavlik, "Training Knowledge-Based Neural Networks To Recognize Genes In DNA Sequences", Neurips, 1990.

[85] R. Maclin, J. W. Shavlik, "Refining Domain Theories Expressed As Finite-State Automata", Elsevier, 1991.
[86] D. Collins, S. Ronen, M. Leach et al., "KBANN's for Classification of Normal Breast 31PM Based on Hormone–Dependent Changes During the Menstrual Cycle", Sussex, 1998.
[87] G. Serpen, D. K. Tekkedil, M. Orra, "A Knowledge-Based Artificial Neural Network Classifier For Pulmonary Embolism Diagnosis", Elsevier, 2007.
[88] N. Fortelny, C. Bock, "Knowledge-Primed Neural Networks Enable Biologically Interpretable Deep Learning On Single-Cell Sequencing Data", Springer, 2020.
[89] N. Holzscheck, C. Falckenhayn, J. Söhle et al., "Modeling Transcriptomic Age Using Knowledge-Primed Artificial Neural Networks", nature, 2021.
[90] P. Gundogdu, C. Loucera, I. Alamo-Alvarez et al., "Integrating Pathway Knowledge With Deep Neural Networks To Reduce The Dimensionality In Single-Cell Rna-Seq Data", IDUS, 2022.
[91] A. Galassi, M. Lippi, P. Torroni, "Investigating Logic Tensor Networks For Neural-Symbolic Argument Mining", AAAI, 2021.
[92] J. G. Diaz Ochoa, L. Maier, O. Csiszar et al., "Bayesian Logical Neural Networks For Human-Centered Applications In Medicine", Frontiers, 2021.
[93] T. Kang, A. Turfah, J. Kim et al., "A Neuro-Symbolic Method For Understanding Free-Text Medical Evidence", Oxford Academic, 2021.
[94] M. Alshahrani, M. A. Khan, O. Maddouri et al., "Neuro-Symbolic Representation Learning On Biological Knowledge Graphs", Oxford Academic, 2017.
[95] M. Kulmanov, F. Z. Smaili, X. Gao et al., "Semantic Similarity And Machine Learning With Ontologies", Springer, 2022.
[96] C. Pesquita, "Towards Semantic Integration For Explainable Artificial Intelligence In The Biomedical Domain", ScitePress, 2021.
[97] M. C. Primo, F. M. Secondo, S. Mazzacane, G. Pagliarini, G. Sciavicco, "Statistical And Symbolic Neuroaesthetics Rules Extraction From EEG Signals", Springer, 2022.
[98] O. Fontenla-Romero, B. Guijarro-Berdiñas, A. Alonso-Betanzos, "Symbolic, Neural And Neuro-Fuzzy Approaches To Pattern Recognition In Cardiotocograms", Springer, 2002.
[99] G. Konstantinopoulou, K. Kovas, I. Hatzilygeroudis, J. Prentzas, "An Approach using Certainty Factor Rules for Aphasia Diagnosis", IEEE Xplore, 2019.
[100] M. Sordo, H. Buxton, D. Watson, "KBANNs and the Classification of 31P MRS of Malignant Mammary Tissue", IEEE Xplore, 1999.
[101] N. Singh, P. Vayer, S. Tanwar et al., "Drug discovery and development: introduction to the general public and patient groups", Frontiers, 2023.
[102] N. Dobberstein, A. Maass, J. Hamaekers, "Llamol: a dynamic multi-conditional generative transformer for de novo molecular design", Springer, 2024.
[103] A. Gaulton, L. J. Bellis, A. P. Bento et al., "ChEMBL: a large-scale bioactivity database for drug discovery", Oxford Academic, 2012.
[104] T. T. Ashburn, K. B. Thor, "Drug repositioning: identifying and developing new uses for existing drugs," nature, 2004.
[105] H. Wang, H. Du, "Knowledge-Enhanced Medical Visual Question Answering: A Survey (Invited Talk Summary)", Springer, 2023.
[106] Z. Lin, D. Zhang, Q. Tao et al., "Medical visual question answering: A survey", ScienceDirect, 2023.
[107] T. Eiter, N. Higuera, J. Oetsch et al., "A Neuro-Symbolic ASP Pipeline for Visual Question Answering", Cambridge University Press, 2022.
[108] J. Moon, "Symmetric Graph-Based Visual Question Answering Using Neuro-Symbolic Approach", MDPI, 2023.
[109] P. Johnston, "A Neuro-Symbolic Incremental Learner Model for the Visual Question Answering Task", University of Stirling, 2023.
[110] T. Eiter, N. N. Higuera, J. Oetsch et al., "A Confidence-Based Interface for Neuro-Symbolic Visual Question Answering", TU Wien, 2022.
[111] D. Xue, S. Qian, C. Xu, "Integrating Neural-Symbolic Reasoning With Variational Causal Inference Network for Explanatory Visual Question Answering", IEEE Xplore, 2024.
[112] T. Eiter, T. Geibinger, N. Higuera et al., "A Logic-based Approach to Contrastive Explainability for Neurosymbolic Visual Question Answering", ACM, 2023.
[113] T. Eiter, N. Higuera Ruiz, J. Oetsch et al., "A Modular Neurosymbolic Approach for Visual Graph Question Answering", CEUR-WS, 2023.
[114] Y. Bao, T. Xing, X. Chen, "Confidence-Based Interactable Neural-Symbolic Visual Question Answering", Elsevier, 2024.
[115] A. Dahlgren, S. Dan, "Compositional Generalization in Neuro-Symbolic Visual Question Answering", Typeset, 2023.
[116] P. Johnston, K. Nogueira, K. Swingler et al., "NS-IL: Neuro-Symbolic Visual Question Answering using Incrementally Learnt, Independent Probabilistic Models for Small Sample Sizes," IEEE Xplore, 2023.
[117] K. Sanders, N. Weir, B. Van Durme, "TV-TREES: Multimodal Entailment Trees for Neuro-Symbolic Video Reasoning," arXiv, 2024.
[118] Z. Lu, I. Afridi, H. J. Kang, I. Ruchkin, X. Zheng, "Surveying Neuro-Symbolic Approaches for Reliable Artificial Intelligence of Things," Springer, 2024.
[119] M. Choi, H. Goel, M. Omama et al., "Neuro-Symbolic Video Search," arXiv, 2024.
[120] A. Abdessaied, M. Bâce, A. Bulling, "Neuro-Symbolic Visual Dialog," arXiv, 2022.
[121] J. Hsu, J. Mao, J. Wu, "Ns3d: Neuro-Symbolic Grounding of 3D Objects and Relations," arXiv, 2023.
[122] M. J. Khan, J. Breslin, E. Curry, "NeuSyRE: Neuro-Symbolic Visual Understanding and Reasoning Framework Based on Scene Graph Enrichment," IOS Press, 2023.
[123] J. Park, S.-J. Bu, S.-B. Cho, "A Neuro-Symbolic AI System for Visual Question Answering in Pedestrian Video Sequences," Springer, 2022.
[124] A. Mileo, "Towards a Neuro-Symbolic Cycle for Human-Centered Explainability," IOS Press.
[125] H. Singh, P. Garg, M. Gupta et al., "Weakly Supervised Neuro-Symbolic Image Manipulation via Multi-Hop Complex Instructions," OpenReview, 2023.
[126] M. J. Khan, F. Ilievski, J. G. Breslin et al., "A Survey of Neurosymbolic Visual Reasoning with Scene Graphs and Common Sense Knowledge," IOS Press.
[127] D. J. Edwards, "A Functional Contextual, Observer-Centric, Quantum Mechanical, and Neuro-Symbolic Approach to Solving the Alignment Problem of Artificial General Intelligence: Safe AI," Frontiers, 2024.
[128] L. Liang, G. Sun, J. Qiu et al., "Neural-Symbolic VideoQA: Learning Compositional Spatio-Temporal Reasoning for Real-World Video Question Answering," arXiv, 2024.
[129] A. Mishra, M. S. Soumitri, V. N. Rajendiran, "Learning Representations from Explainable and Connectionist Approaches for Visual Question Answering," IEEE Xplore, 2024.
[130] W. Zhu, J. Thomason, R. Jia, "Generalization Differences between End-to-End and Neuro-Symbolic Vision-Language Reasoning Systems," arXiv, 2022.
[131] J. Gao, A. Blair, M. Pagnucco, "Explainable Visual Question Answering via Hybrid Neural-Logical Reasoning," IEEE Xplore, 2024.
[132] L. Mitchener, D. Tuckey, M. Crosby et al., "Detect, Understand, Act: A Neuro-Symbolic Hierarchical Reinforcement Learning Framework," Springer, 2022.

[133] J. J. Bauer, T. Eiter, N. Higuera Ruiz et al., "Neuro-Symbolic Visual Graph Question Answering with LLMs for Language Parsing," TU Wien, 2023.
[134] J. Kwon, J. Tenenbaum, S. Levine, "Neuro-Symbolic Models of Human Moral Judgment," 2024.
[135] J. Sha, H. Shindo, K. Kersting et al., "Neuro-Symbolic Predicate Invention: Learning Relational Concepts from Visual Scenes," IOS Press, 2024.
[136] A. K. Kovalev, M. Shaban, E. Osipov et al., "Vector Semiotic Model for Visual Question Answering," Elsevier, 2022.
[137] D. Lindström, A. Dahlgren, "Learning, Reasoning, and Compositional Generalisation in Multimodal Language Models," 2024.
[138] J. Renkhoff, K. Feng, M. Meier-Doernberg et al., "A Survey on Verification and Validation, Testing, and Evaluations of Neurosymbolic Artificial Intelligence," IEEE Xplore, 2024.
[139] A. Jain, A. R. Kondapally, K. Yamada et al., "A Neuro-Symbolic Approach for Multimodal Reference Expression Comprehension," arXiv, 2023.
[140] J. Gao, A. Blair, M. Pagnucco, "A Symbolic-Neural Reasoning Model for Visual Question Answering," IEEE Xplore, 2023.
[141] P. Vakharia, A. Kufeldt, M. Meyers et al., "ProSLM: A Prolog Synergized Language Model for Explainable Domain-Specific Knowledge-Based Question Answering," Springer, 2024.
[142] Y. Wang, Z. Tu, Y. Xiang et al., "Rapid Image Labeling via Neuro-Symbolic Learning," ACM, 2023.
[143] M. Hersche, F. di Stefano, T. Hofmann et al., "Probabilistic Abduction for Visual Abstract Reasoning via Learning Rules in Vector-Symbolic Architectures," arXiv, 2024.
[144] G. Burghouts, F. Hillerström, E. Walraven et al., "Open-World Visual Reasoning by a Neuro-Symbolic Program of Zero-Shot Symbols," arXiv, 2024.
[145] Z. Li, Y. Yao, T. Chen et al., "Softened Symbol Grounding for Neuro-Symbolic Systems," arXiv, 2024.
[146] S. S. Abraham, M. Alirezaie, L. De Raedt, "CLEVR-POC: Reasoning-Intensive Visual Question Answering in Partially Observable Environments," arXiv, 2024.
[147] X. Yang, F. Liu, G. Lin, "Neural Logic Vision Language Explainer," IEEE Xplore, 2023.
[148] R. Feinman, "Generative Neuro-Symbolic Models of Concept Learning," 2023.
[149] Z. Wan, C.-K. Liu, H. Yang et al., "Towards Cognitive AI Systems: A Survey and Prospective on Neuro-Symbolic AI," arXiv, 2024.
[150] S. Das, B. Zhou, "Hybrid Neuro-Symbolic Reasoning Based on Multimodal Fusion," 2023.
[151] Y. Bao, T. Xing, X. Chen, "A Confidence-Based Multipath Neural-Symbolic Approach for Visual Question Answering," IJCAI, 2023.
[152] J. Levine, "Methods for Generating Visual Programs with Optimizable Vision Models," 2024.
[153] M. Endo, J. Hsu, J. Li et al., "Motion Question Answering via Modular Motion Programs," MLR Press, 2023.
[154] T. Gupta, A. Kembhavi, "Visual Programming: Compositional Visual Reasoning Without Training," IEEE Xplore, 2023.
[155] A. R. Jacob, "Neuro-Argumentative Machine Learning with Images," 2024.
[156] D. Reich, T. Schultz, "On the Role of Visual Grounding in VQA," arXiv, 2024.
[157] C. Feng, J. Hsu, W. Liu et al., "Naturally Supervised 3D Visual Grounding with Language-Regularized Concept Learners," IEEE Xplore, 2024.
[158] K. Yi, J. Wu, C. Gan et al., "Neural-Symbolic VQA: Disentangling Reasoning from Vision and Language Understanding," ACM, 2018.
[159] J. Johnson, B. Hariharan, L. van der Maaten et al., "CLEVR: A Diagnostic Dataset for Compositional Language and Elementary Visual Reasoning," Stanford University, 2017.
[160] J. Mao, C. Gan, P. Kohli et al., "The Neuro-Symbolic Concept Learner: Interpreting Scenes, Words, and Sentences From Natural Supervision," arXiv, 2019.
[161] S. Antol, A. Agrawal, J. Lu et al., "VQA: Visual Question Answering," IEEE Xplore, 2015.
[162] D. A. Hudson, C. D. Manning, "GQA: A New Dataset for Real-World Visual Reasoning and Compositional Question Answering," arXiv, 2019.
[163] J. Andreas, M. Rohrbach, T. Darrell et al., "Neural Module Networks," arXiv, 2017.
[164] R. Hu, J. Andreas, M. Rohrbach et al., "Learning to Reason: End-to-End Module Networks for Visual Question Answering," ICCV, 2017.
[165] D. A. Hudson, C. D. Manning, "Learning by Abstraction: The Neural State Machine," ACM, 2019.
[166] N. Díaz-Rodríguez, A. Lamas, J. Sanchez et al., "EXplainable Neural-Symbolic Learning (X-NeSyL) Methodology to Fuse Deep Learning Representations with Expert Knowledge Graphs: The MonuMAI Cultural Heritage Use Case," Elsevier, 2021.
[167] A. Karim, M. Lee, T. Balle et al., "CardioTox Net: A Robust Predictor for hERG Channel Blockade Based on Deep Learning Meta-Feature Ensembles," Springer, 2021.
[168] T. Liu, L. Hwang, S. K. Burley et al., "BindingDB in 2024: A FAIR Knowledgebase of Protein-Small Molecule Binding Data," ChemRxiv, 2024.
[169] S. Kim, J. Chen, T. Cheng et al., "PubChem 2023 Update," Oxford Academic, 2022.
[170] S. Wang, H. Sun, H. Liu et al., "ADMET Evaluation in Drug Discovery. 16. Predicting hERG Blockers by Combining Multiple Pharmacophores and Machine Learning Approaches," ACS Publications, 2016.
[171] S. D. Harding, J. F. Armstrong, E. Faccenda et al., "The IUPHAR/BPS Guide to PHARMACOLOGY in 2024," Oxford Academic, 2023.
[172] J. Gilmer, S. S. Schoenholz, P. F. Riley et al., "Neural Message Passing for Quantum Chemistry," arXiv, 2017.
[173] L. Breiman, "Random Forests," Springer, 2001.
[174] G. Xiong, Z. Wu, J. Yi et al., "ADMETlab 2.0: An Integrated Online Platform for Accurate and Comprehensive Predictions of ADMET Properties," Oxford Academic, 2021.
[175] J. H. Friedman, "Greedy Function Approximation: A Gradient Boosting Machine," Stanford University, 2001.
[176] Y. Qiang, Y. Yang, X. Zhang et al., "Tensor Composition Net for Visual Relationship Prediction," arXiv, 2022.
[177] T. Mikolov, K. Chen, G. Corrado, and J. Dean, "Efficient estimation of word representations in vector space," arXiv, 2013.
[178] J. Pennington, R. Socher, and C. D. Manning, "Glove: Global vectors for word representation," 2014
[179] T. Brown, B. Mann, N. Ryder et al., "Language models are few-shot learners," 2020
[180] D. Silver, A. Huang, C. J. Maddison et al., "Mastering the game of go with deep neural networks and tree search," nature, 2016.
[181] X. Chen, C. Liang, A. W. Yu et al., "Compositional generalization via neural-symbolic stack machines," 2020
[182] R. Dang-Nhu, "Plans: Neuro-symbolic program learning from videos," 2020
[183] T. Gupta and A. Kembhavi, "Visual programming: Compositional visual reasoning without training," IEEE, 2023
[184] Y. Shen, K. Song, X. Tan et al., "Hugginggpt: Solving ai tasks with chatgpt and its friends in huggingface," 2023.
[185] D. Sur´ıs, S. Menon, and C. Vondrick, "Vipergpt: Visual inference via python execution for reasoning," IEEE, 2023.
[186] W.-Z. Dai, Q. Xu, Y. Yu et al., "Bridging machine learning and logical reasoning by abductive learning," 2019
[187] G. Lample and F. Charton, "Deep learning for symbolic mathematics," 2019.

[188] R. Evans and E. Grefenstette, “Learning explanatory rules from noisy data,” 2018.
[189] P. Hohenecker and T. Lukasiewicz, “Ontology reasoning with deep neural networks,” 2020.
[190] Y. Liang and G. Van den Broeck, “Learning logistic circuits,”AAAI, 2019.
[191] T. Demeester, T. Rocktaschel, and S. Riedel, “Lifted rule injection ¨ for relation embeddings,” 2016.
[192] Z. Hu, X. Ma, Z. Liu et al., “Harnessing deep neural networks with logic rules,” ACL, 2016.
[193] R. Stewart and S. Ermon, “Label-free supervision of neural networks with physics and domain knowledge,” AAAI, 2017.
[194] N. Muralidhar, M. R. Islam, M. Marwah, A. Karpatne, and N. Ramakrishnan, “Incorporating prior domain knowledge into deep neural networks,” IEEE, 2018.
[195] P.-W. Wang, P. Donti, B. Wilder, and Z. Kolter, “Satnet: Bridging deep learning and logical reasoning using a differentiable satisfiability solver,” 2019.
[196] J. Cai, R. Shin, and D. Song, “Making neural programming architectures generalize via recursion,” 2017.
[197] L. Li, W. Wang, and Y. Yang, “Logicseg: Parsing visual semantics with neural logic learning and reasoning,” in ICCV, 2023.
[198] Sasha B. Ebrahimi, Devleena Samanta, “Engineering protein-based therapeutics through structural and chemical design,” nature, 2023
[199] Qingliang Li, Luhua Lai, Prediction of potential drug targets based on simple sequence
Properties,” BMC Bioinformatics, 2007
[200] Sarfaraz K. Niaz, Zamara Mariam, “Reinventing Therapeutic Proteins: Mining a Treasure of New Therapies,” MDPI, 2023
[201] Bing-Xue Du, Yuan Qin, Yan-Feng Jiang et al., “Compound–protein interaction prediction by deep learning: Databases, descriptors and models,” Science Direct, 2022
[202] Delower Hossain, Jake Y. Chen, Fuad Al Abir, “hERG-LTN: A New Paradigm in hERG Cardiotoxicity Assessment Using Neuro-Symbolic and Generative AI Embedding (MegaMolBART, Llama3.2, Gemini, DeepSeek) Approach.” bioRxive, 2025, doi: https://doi.org/10.1101/2025.02.17.638731

## APPENDIX A: A VARITY OF NESY DOMAIN AND APPLICATIONS PRESENTED AS BELOW.

| Domain Of Application | Model | Reference |
|---|---|---|
| Healthcare | KBANN | GG Towell et al., 1994 |
| | LTN | S Badreddine et al., 2022 |
| | Deepinfusion | I. Abdullah., 2023 |
| | ExplainDR | SI Jang et al ., 2021 |
| | NeSyL | N Díaz-Rodríguez et al., 2022 |
| | BioBert+Mt Attention | T. Kang et al ., 2021 |
| Cyber Security | GA-NeSy | KW Park et al ., 2021 |
| | Approach | A Piplai et al ., 2023 |
| Recommendation System | CARS | D Bouneffouf et al ., 2014 |
| | CTR | D Bouneffouf et al ., 2012 |
| | KARF | G Spillo et al ., 2021 |
| | NLQ4Rec | M Wu et al 2024 |
| | LPFR | X Tong et al., 2024 |
| Robotics/Automation | Approach | Y Zhang et al 2023 |
| | Approach | A Gomaa et al., 2023 |
| | Sophia | D Hanson et al., 2020 |
| Smart City | Approach | G Morel et al ., 2021 |
| Information Retrieval | Approach | L Dietz et al., 2023 |
| NLP | Review | K Hamilton et al., 2022 |
| | NMN | J Andreas et al., 2015 |
| | NLProlog | L Weber et al., 2019 |
| | Approach | K Gupta et al., 2021 |
| | PIGLeT | R Zellers et al., 2021 |
| | LNN-El | H Jiang et al., 2021 |
| VQA / Computer Vision | NS-VQA | K Yi et al., 2019 |
| | NS-CL | J Mao et al., 2019 |
| | Approach | A Mishra et al., 2024 |
| | XNM | J Shi et al., 2019 |
| | NS-IL | P Johnston et al., 2023 |
| | Approach | J Park et al., 2022 |
| Generative AI | TITScore | P Ji et al., 2024 |
| | Approach | N karacapilidis et al., 2024 |
| | LIDA | P Agrawal et al., 2024 |
| | Genome | Z Chen et al., 2023 |
| | ContextGPT | L Arrotta et al., 2024 |

**APPENDIX: B**

**Table:** Acronyms and Abbreviations

| | |
|---|---|
| AAAI | Association for the Advancement of Artificial Intelligence |
| ACL | Association for Computational Linguistics |
| AI | Artificial Intelligence |
| CBR | Case-based reasoning |
| CNN | Convolutional Neural Network |
| DL | Deep Learning |
| GCNN | Graph Convolutional Neural Network |
| GPT3 | Third-generation Generative Pre-trained Transformer |
| IJCAI | International Joint Conference on Artificial Intelligence |
| KG | Knowledge Graphs |
| LNN | Logical Neural Networks |
| LLM | Large Language Models |
| LTN | Logic Tensor Network |
| ML | Machine Learning |
| MLP | Multilayer Perceptron |
| NeSy | Neuro-Symbolic AI |
| NLP | Natural Language Processing |
| NS-CL | Neuro-Symbolic Concept Learner |
| NN | Neural Network |
| OOD | Out-of-distribution |
| ProbLog | Probabilistic Logic Programming |
| SOTA | State of the Art |
| ACM | Association for Computing Machinery |
| NeurIPS | Conference on Neural Information Processing Systems |
| IDUS | Institutional Repository of the University of Seville |
| IJCLR | International Journal of Collaborative Learning Research |
| HEALTHINF | International Conference on Health Informatics |
| ScitePress | Science and Technology Publications |
| IEEE | Institute of Electrical and Electronics Engineers |
| RSC | Royal Society of Chemistry |
| AIP | American Institute of Physics |
| RQ | Research Question |

## APPENDIX: C

**Table: A List of Neuro-Symbolic Models GitHub Repository**

| | Model/Algorithm | GitHub Link | Dataset | Data Type | Author |
|---|---|---|---|---|---|
| 1. | **KBANN/KBNN** | https://github.com/spacewalk01/deep-logic-learning | Promoter & splice-junction datasets | Tabular | Geoffrey G. Towell |
| 2. | **NEFCLASS-J** | https://github.com/dhecloud/NEFCLASS | Wisconsin Breast Cancer" (WBC) | Tabular | D. D. Nauck |
| 3. | **KPNN** | https://github.com/epigen/KPNN | RNA-seq dataset | Tabular | Nikolaus Fortelny |
| 4. | **SHERLOCK** | https://github.com/sherlock-project/sherlock | UCI Machine Learning Repository | Tabular | Ekaterina |
| 5. | **NMN** | https://github.com/jacob&reas/nmn2 | VQA | Image & Text | Jacob &reas |
| 6. | **NeuralLP** | https://github.com/fanyangxyz/Neural-LP | WikiMovies, WordNet18, Freebase15K | Tabular | Yang |
| 7. | **DeepProbLog** | https://github.com/ML-KULeuven/deepproblog | MNIST | Image | Robin Manhaeve |
| 8. | **CommonsenseQA** | https://github.com/jonathanherzig/commonsenseqa | DREAM/CommonsenseQA | Text | Kaixin Ma |
| 9. | **ANSNA** | https://github.com/patham9/ANSNA | Giuseppe Marra | Text | Patrick Hammer |
| 10. | **NLProlog** | https://github.com/leonweber/nlprolog | MEDHOP, WIKIHOP | Image | Leon Weber |
| 11. | **NS-CL** | https://github.com/kexinyi/ns-vqa | CLEVR | Tabular | Jiayuan Mao |
| 12. | **NMLN** | https://github.com/GiuseppeMarra/nmln | Nations, Kinship & UMLS, SMOKER | Text | Giuseppe Marra |
| 13. | **CORGI** | https://github.com/ForoughA/CORGI | **Commonsense Reasoning** | Text | Forough Arabshahi1 |
| 14. | **DFOL-VQA** | https://github.com/microsoft/DFOL-VQA | GQA | Tabular | Saeed Amizadeh |
| 15. | **GNS-model** | https://github.com/rfeinman/GNS-Modeling | Omniglot dataset | Text | Reuben Feinman |
| 16. | **Hybrid-CEP** | https://github.com/KyraStyl/Hybrid_CEP_Engine | Urban Sounds 8 | Audio | Marc Roig Vilamala, Harrison Taylor |
| 17. | **R3-Transformer** | https://github.com/hassanhub/R3Transformer | YouCook2 | Tabular | Hassan Akbari |
| 18. | **KBQA** | https://github.com/svakulenk0/KBQA | QALD - 9/ LC-QuAD 1.0 | Text | Pavan Kapanipathi |
| 19. | **NeSy XIL** | https://github.com/ml-research/NeSyXIL | CLEVR-Hans/ColorMNIST | Image | Wolfgang Stammer |
| 20. | **LNN** | https://github.com/IBM/LNN | Smokers & friends dataset-LTN , & LUBM | Tabular | Ryan Riegel |
| 21. | **NeurASP** | https://github.com/azreasoners/NeurASP | [Xu et al., 2018] | Tabular | Zhun Yang |
| 22. | **MD-informed** | https://github.com/Tian312/MD-Attention | COVID-19 data | Tabular | Tian Kang |
| 23. | **GABL** | https://github.com/AbductiveLearning/GABL | Synthesized dataset | Tabular | Le-Wen Cai |
| 24. | **RWFN** | https://github.com/jyhong0304/SII | PASCAL-Part-dataset, ontologies (WordNet) | object detection image | Jinyung Hong |
| 25. | **autoBOT** | https://github.com/aiden/autobot | ---- | Text | Blaž Škrlj |
| 26. | **DPL** | https://github.com/travis-ci/dpl | IMDb, Yahoo dataset | Tabular | Hoifung Poon |
| 27. | **RNNLOGIC** | https://github.com/DeepGraphLearning/RNNLogic | FB15k-237, WN18RR | Tabular | Meng Qu∗ |
| 28. | **COMET** | https://github.com/dotnet/Comet | SocialIQa, StoryCS | Tabular | Antoine Bosselut |
| 29. | **NSNnet** | https://github.com/GuillaumeVW/NSNet | MNIST | Image | Ananye Agarwal |
| 30. | **Faster-LTN** | https://gitlab.com/grains2/Faster-LTN | PASCAL VOC, PASCAL PART | Object detection image | Francesco Manigrasso1 |
| 31. | **EduCe** | https://github.com/magiclen/educe | FB15K-237, Constant, Family-gender, UMLS | Image | Anonymous ACL |
| 32. | **CTP** | https://github.com/microsoft/CtP | CLUTRR | Tabular | Pasquale Minervini |
| 33. | **NEUROSPF** | https://github.com/neurospf/neurospf | MNIST | Image | Muhammad Usman |
| 34. | **NESTER** | https://github.com/tapnair/NESTER | ICFHR'14 CROHME | Image | Paolo Dragone |
| 35. | **NSFR** | https://github.com/ml-research/nsfr | 2D K&insky patterns & 3D CLEVR-Hans | Image & Text | Hikaru Shindo |
| 36. | **HRI** | https://github.com/Alvearie/HRI | Visual Genome | Image | Claire Glanois |
| 37. | **PIGLeT** | https://github.com/Dervall/Piglet | EXIST | -- | Rowan Zellers |
| 38. | **Pix2rule** | https://github.com/nuric/pix2rule | Subgraph set isomorphism | -- | Nuri Cingillioglu |
| 39. | **Latplan** | https://github.com/guicho271828/latplan | Twisted LightsOut | -- | Masataro Asai |
| 40. | **NS-Dial** | https://github.com/shiquanyang/NS-Dial | MultiWOZ 2.1, SMD | Text | Shiquan Yang |
| 41. | **DNR** | https://github.com/SkBlaz/DNR | BITCOIN | Tabular | Blaˇz Skrlj |
| 42. | **NS-ICP** | https://github.com/EdmundYanJ/NS-ICF | ,ml-1m, Taobao | Tabular | Wei Zhang |
| 43. | **RETOMATON** | https://github.com/neulab/retomaton | WIKITEXT-103, Law-MT | Text | Uri Alon |